\documentclass[runningheads]{llncs}

\usepackage{eccv}

\usepackage{eccvabbrv}

\usepackage{graphicx}
\usepackage{booktabs}

\usepackage[accsupp]{axessibility}  

\usepackage{amsmath,amsfonts,bm}

\def\eqref#1{equation~\ref{#1}}

\def\1{\bm{1}}

\DeclareMathAlphabet{\mathsfit}{\encodingdefault}{\sfdefault}{m}{sl}
\SetMathAlphabet{\mathsfit}{bold}{\encodingdefault}{\sfdefault}{bx}{n}

\def\gD{{\mathcal{D}}}

\def\gN{{\mathcal{N}}}

\usepackage{multirow}
\usepackage{pifont}
\usepackage{wrapfig,lipsum,booktabs}

\usepackage{listings}
\definecolor{codegreen}{rgb}{0,0.6,0}
\definecolor{codegray}{rgb}{0.5,0.5,0.5}
\definecolor{codepurple}{rgb}{0.58,0,0.82}
\definecolor{backcolour}{rgb}{0.95,0.95,0.92}
\lstdefinestyle{mystyle}{
    backgroundcolor=\color{backcolour},
    commentstyle=\color{codegreen},
    keywordstyle=\color{magenta},
    numberstyle=\tiny\color{codegray},
    stringstyle=\color{codepurple},
    basicstyle=\ttfamily\footnotesize,
    breakatwhitespace=false,         
    breaklines=true,                 
    captionpos=b,                    
    keepspaces=true,                 
    numbers=left,                    
    numbersep=5pt,                  
    showspaces=false,                
    showstringspaces=false,
    showtabs=false,                  
    tabsize=2
}
\newcommand{\ours}{DataDream}
\newcommand{\oursdset}{$\text{DataDream}_{\text{dset}}$}
\newcommand{\ourscls}{$\text{DataDream}_{\text{cls}}$}
\newcommand{\oursbase}{Real-finetune}
\newcommand{\vmark}{\ding{51}}

\newcommand{\myparagraph}[1]{\noindent{\bf{#1}}}

\usepackage{hyperref}

\usepackage{orcidlink}

\begin{document}

    \title{DataDream: Few-shot Guided Dataset Generation} 


\author{
Jae Myung Kim$^{1,2,3,*}$
\;\;\;\; Jessica Bader$^{2,3,4,*}$
\;\;\;\; Stephan Alaniz$^{2,3}$ \\ 
Cordelia Schmid$^{5}$
\;\;\;\; Zeynep Akata$^{2,3,4}$
}

\authorrunning{J.~Kim et al.}

\institute{
\small{$^{1}$University of Tübingen}
\;\;\;\; \small{$^{2}$Helmholtz Munich}
\;\;\;\; \small{$^{3}$MCML}
\;\;\;\; \small{$^{4}$TUM} \\ 
\small{$^{5}$Inria, Ecole normale sup\'erieure, CNRS, PSL Research University}\\
~\\
\small{$^*$equal contribution}
}

\maketitle

\begin{abstract}
    While text-to-image diffusion models have been shown to achieve state-of-the-art results in image synthesis, they have yet to prove their effectiveness in downstream applications. Previous work has proposed to generate data for image classifier training given limited real data access. However, these methods struggle to generate in-distribution images or depict fine-grained features, thereby hindering the generalization of classification models trained on synthetic datasets. We propose DataDream, a framework for synthesizing classification datasets that more faithfully represents the real data distribution when guided by few-shot examples of the target classes. DataDream fine-tunes LoRA weights for the image generation model on the few real images before generating the training data using the adapted model. We then fine-tune LoRA weights for CLIP using the synthetic data to improve downstream image classification over previous approaches on a large variety of datasets. We demonstrate the efficacy of DataDream through extensive experiments, surpassing state-of-the-art classification accuracy with few-shot data across 7 out of 10 datasets, while being competitive on the other 3. Additionally, we provide insights into the impact of various factors, such as the number of real-shot and generated images as well as the fine-tuning compute on model performance. 
    The code is available at \href{https://github.com/ExplainableML/DataDream}{https://github.com/ExplainableML/DataDream}.

\end{abstract}

\section{Introduction}
\label{sec:intro}

\begin{figure}[]
    \centering
    \begin{minipage}{0.49\textwidth}
        \centering
        \includegraphics[width=\linewidth]{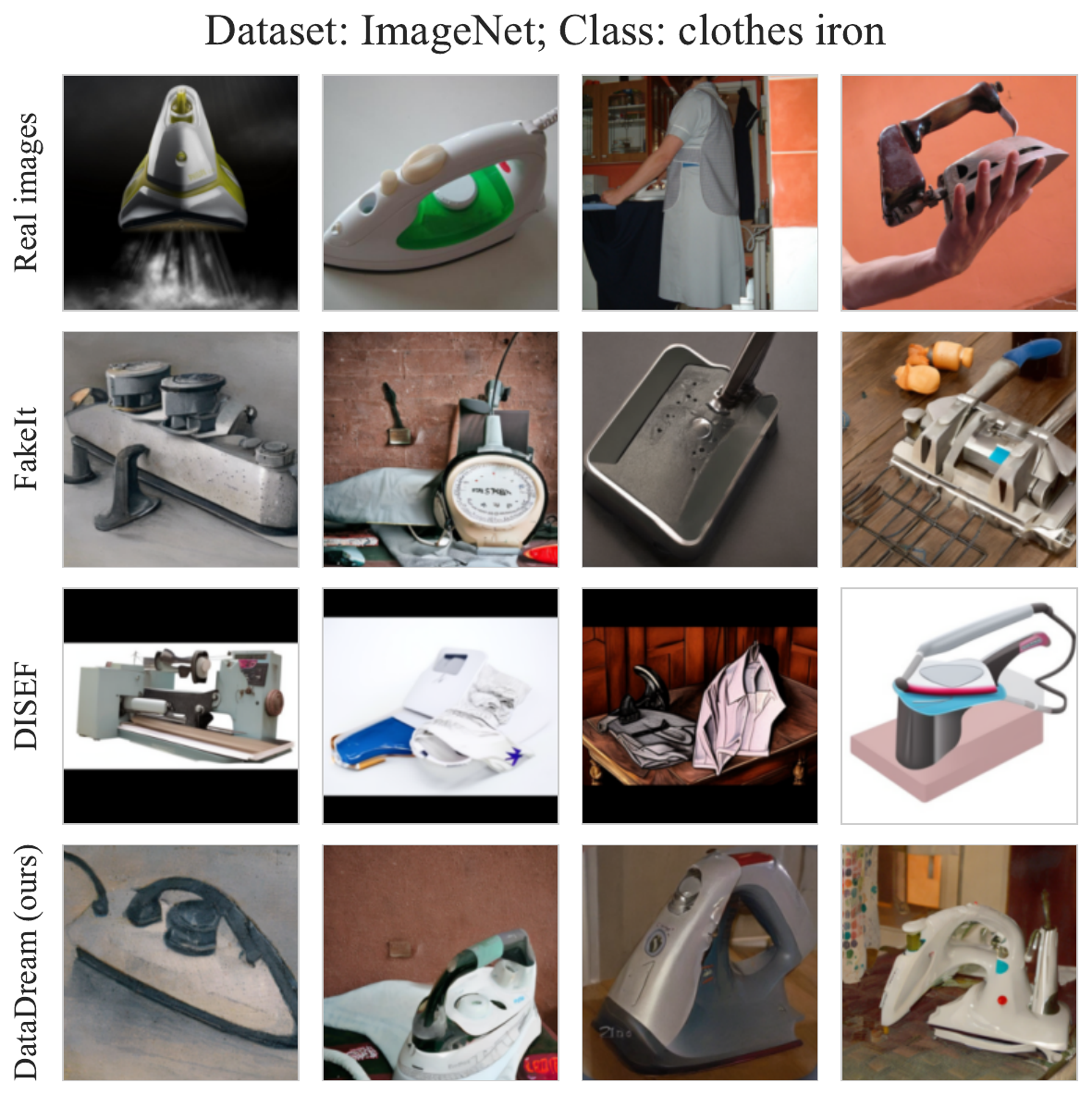}
    \end{minipage}\hfill
    \begin{minipage}{0.49\textwidth}
        \centering
        \includegraphics[width=\linewidth]{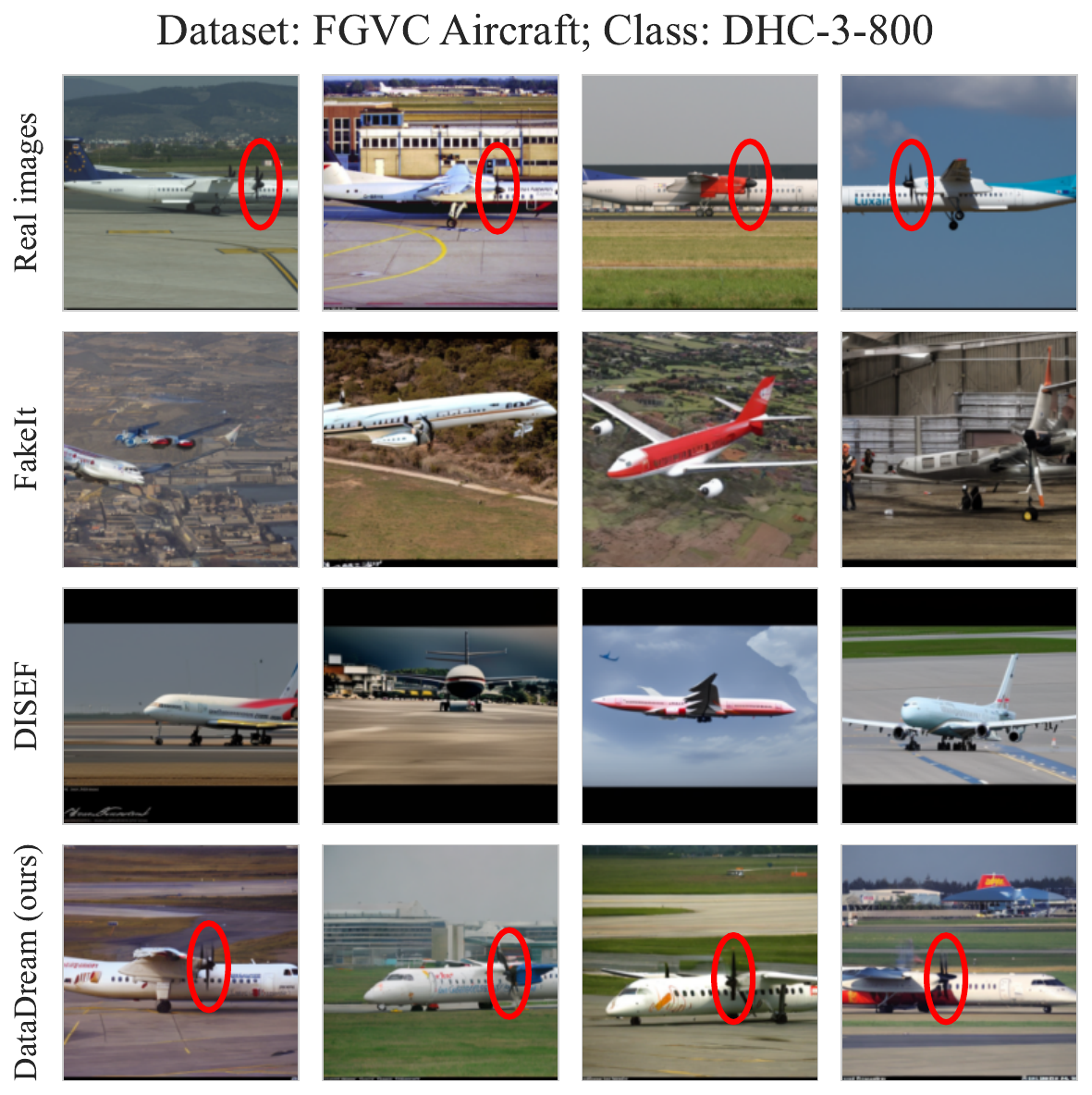}
    \end{minipage}
    \caption{
        \textbf{Synthetic images comparison.} The previous methods for synthesizing training data sometimes misunderstand the class name due to its ambiguity (FakeIt~\cite{fakeit} confuses the clothes iron with the metal iron) or fail to capture fine-grained features (DISEF~\cite{disef} generated images lack the propeller in front of the wings in the DHC-3-800 aircraft, a red circle indicates the propeller). Meanwhile, our method accurately generates images of the class of interest and captures fine-grained details. 
    }    
    \label{fig:teaser}
\end{figure}

The emergence of text-to-image generative models, such as Stable Diffusion~\cite{stable_diffusion}, not only enables us to create photo-realistic synthetic images, but it also presents opportunities to enhance downstream tasks. One potential application lies in training or fine-tuning task-specific models on synthetic data. This is shown to be particularly useful in domains where access to real data is limited~\cite{issynth,disef,lin2023explore,tian2024stablerep}, as generative models offer a cost-effective means of generating large amounts of training data. In this paper, we study the impact of synthetic training data on image classification tasks in low-shot settings, i.e. where we have access to a few images per class, but the collection of an entire dataset would be prohibitively expensive.

Previous research has primarily focused on using the class names of a given dataset~\cite{fakeit,shipard2023diversity,yu2023diversify,issynth} to inform the data generation process. Concretely, they generated images with text-to-image diffusion models, using the class names as conditional input. To better guide the model to generate accurate depictions of the target object, they incorporated textual descriptions of each class to the prompt, sourced from language models~\cite{yu2023diversify,issynth} or human-annotated class descriptions~\cite{fakeit}. While intuitive, these methods lead to some generated images lacking the object of interest. For instance, while the real images for the class name \texttt{"clothes iron"} from the ImageNet~\cite{imagenet} dataset display the appliance for ironing clothes, the images generated by FakeIt~\cite{fakeit} mostly depict iron as the metal or arbitrary objects made thereof (Figure~\ref{fig:teaser}, left). This occurs when the generative model misunderstands class name ambiguities or rare classes. Such misalignment between the real and synthetic images limits the generated images' informational value for image classification and hinders performance gains. 

To bridge the gap between real and synthetic images, real images can better inform the generative model about the characteristics of the real data distribution~\cite{issynth,dunlap2023diversify,disef,azizi2023synthetic,zhou2023training}. For instance, the concurrently developed DISEF~\cite{disef} method uses few-shot samples as conditional input to the pre-trained diffusion model by starting from a partially noised real image when generating the synthetic dataset. It additionally uses a pre-trained captioning model to diversify the text-to-image prompt. While this approach improves the alignment of real and synthetic data distributions, it sometimes falls short of capturing fine-grained features. For example, while the real images for the class name \texttt{"DHC-3-800"} in the Aircraft~\cite{aircraft} dataset include a propeller in front of the wings, the synthetic images by DISEF lack this detail (Figure~\ref{fig:teaser}, right). Accurately representing class-discriminative features can be critical for classification tasks, particularly in fine-grained datasets.

In this work, we propose a novel approach, called \textbf{\ours}, aimed at adapting generative models using few-shot real data. Motivated by personalized generative modeling methods~\cite{textual_inv,dreambooth}, in which generation models are fine-tuned with a small set of real images depicting an \textit{identical object}, our method focuses on aligning the generative model to a target dataset which has \textit{multiple classes and diverse objects for each class}. This differs from previous few-shot dataset generation methods such as~\cite{issynth,disef}, which have not explored fine-tuning the generative model. Concretely, we adapt Stable Diffusion~\cite{stable_diffusion} with LoRA~\cite{lora} in two ways: \textbf{\ourscls}, which trains LoRA per class, and \textbf{\oursdset}, which trains a single LoRA for all classes. To the best of our knowledge, we are the first to propose using few-shot data to adapt the generative model for synthetic training data, rather than leveraging the frozen, pre-trained generation model. Following training, we generate images with the same prompt used for fine-tuning \ours, resulting in images depicting the object of interest (e.g. the clothes iron) or fine-grained features (e.g. the propeller of the DHC-3-800 plane) as shown in the last row of Figure~\ref{fig:teaser}.

We demonstrate the effectiveness of \ours\ through extensive experiments, achieve the state of the art across all datasets when using only synthetic data, and achieve the best performance on 7 out of 10 datasets when training with both real few-shot and synthetic data. To understand the effectiveness of our method, we analyze the alignment between real and synthetic data, revealing that our method shows better alignment with the distribution of real data compared to baseline methods. Finally, we explore the scalability of our method by increasing the number of synthetic data points and real samples, showing the potential benefits of larger datasets. To summarize, the contributions of our work are as follows: 
\begin{enumerate}
    \item We introduce \ours, a novel few-shot method which adapts Stable Diffusion to generate better in-distribution images for downstream training that  outperforms state-of-the-art few-shot classification on 7 out of 10 datasets, with the other 3 comparable.
    \item We emphasize the importance of reporting results with only synthetic data. We demonstrate that our method achieves superior performance when training the classifier with solely synthetic data, in some cases outperforming those trained solely with real few-shot images, indicating that our method generates images that glean more insightful information from the few-shot real data.
    \item We study the effectiveness of our method by analyzing the distribution alignment between synthetic data and real data. Under few-shot guidance, synthetic data generated by our method aligns the best with real data.
\end{enumerate}

\section{Related work}
\label{sec:related_work}

Synthetic image generation has made immense progress, now being capable of generating images that even humans may find difficult to distinguish from real images. In the following, we review related work on image generation and training on synthetic data.\\

\myparagraph{Synthetic Image Generation.} 
The suite of image generation models is growing, including Variational Auto-Encoders \cite{kingma2013}, GANs \cite{goodfellow2014}, and Diffusion Models \cite{stable_diffusion}. With their recent popularity, diffusion models such as Stable Diffusion \cite{stable_diffusion}, SDXL \cite{podell2023sdxl}, DALL-E \cite{dalle2,dalle3}, Imagen \cite{saharia2022photorealistic}, GLIDE \cite{nichol2021glide}, and Wuerstchen \cite{pernias2023wuerstchen} have revolutionized text-to-image generation. Diffusion models aim to incrementally de-noise data by modeling the reverse process of a Markov chain progressively adding Gaussian noise to the sample conditioned on text. At test-time, this facilitates the generation of synthetic images from specified text and random noise. These large pre-trained models can be adapted to user specific needs \cite{textual_inv,dreambooth} or better control generation \cite{liu2023more,choi2021ilvr}. Textual inversion \cite{textual_inv} uses a small number of images of a specific object to learn a representational language token which can be used to prompt the frozen generation model to create better images of that object (e.g. a photo of \textit{your} cat, rather than \textit{a} cat). On the other hand, DreamBooth \cite{dreambooth} achieves personalization by fine-tuning the generation model while providing a unique input token with two losses: one to reconstruct the personalized concept, and the other to preserve the original model generations without the unique token.\\

\myparagraph{Training with Synthetic Data.} 
A pool of research has blossomed in its wake, exploring downstream applications; namely: can models be trained on synthetic data? Some works augmented real datasets with synthetic images \cite{dunlap2023diversify, zhou2023training, bansal2023leaving, burg2023image}. Others focused on pre-training on large amounts of synthetic data, followed by fine-tuning on a limited number of real images \cite{tian2024stablerep, hammoud2024synthclip}. Similarly, several works evaluated the effectiveness of training on entirely synthetic datasets \cite{fakeit, hammoud2024synthclip}.

Different tasks have been considered, including classification \cite{issynth, fakeit, shipard2023diversity, zhou2023training, bansal2023leaving, burg2023image}, object detection \cite{lin2023explore}, image generation \cite{alemohammad2023self}, and representation learning \cite{tian2024stablerep}. Attempts have been made to optimize the selection process from large pools of synthetic data, generally by focusing on two primary factors: faithfulness and diversity. Faithfulness has been addressed by CLIP filtering \cite{issynth, dunlap2023diversify, lin2023explore}, including additional class information \cite{fakeit}, and spectral clustering \cite{lin2023explore}. On the other hand, diversity can be increased by lowering the guidance scale \cite{fakeit}, generating a wide variety of natural language prompts with LLMs \cite{issynth, hammoud2024synthclip}, specifying domains \cite{shipard2023diversity, dunlap2023diversify} or backgrounds \cite{fakeit}, and using multiple text prompt templates \cite{burg2023image}. Generally, data collection is considered resource intensive, while generating synthetic data is comparatively inexpensive and can therefore be done at scale; \cite{fakeit} showed that as the number of synthetic images increases, model performance can even surpass that of models trained on a lower fixed number of real images.

Finally, the few-shot setting is seeing increased interest, where the focus lies on leveraging large amounts of synthetic data in conjunction with limited amounts of real data. More than simply pooling the data sources together, we can guide the generation of better synthetic data with real data. In \cite{issynth}, the authors explored two strategies: 1) 
generating images by starting from a partially noised few-shot sample and
2) using the similarity of synthetic image features to real ones to remove low-confidence samples. When adapting the CLIP model using Classifier Tuning \cite{wortsman2022robust} the first strategy works best. Concurrently to our work, Diversified In-domain Synthesis with Efficient Fine-tuning (DISEF) \cite{disef} proposes to create a synthetic augmentation pipeline which leverages few-shots by starting the geration process from a noised real sample (same as~\cite{issynth}), then promotes diversity by denoising it conditioned on the caption from a different real image. The authors apply CLIP filtering to remove synthetic images which would be classified incorrectly and then adapt CLIP as the classifier with LoRA~\cite{lora} on either the few-real shots alone or the combination of few-shots and synthetic data.
In contrast to these methods, we propose to additionally fine-tune the diffusion model with LoRA to obtain a better alignment with the real data distribution.\\

\section{Methodology}
\label{sec:method}

In this section, we start by describing the preliminaries in \S\ref{subsec:preliminaries}, before introducing DataDream in \S\ref{subsec:datadream}. DataDream fine-tunes the text-to-image diffusion model with few-shot data. To measure performance, synthetic images are generated with the adapted model and a classifier trained on both synthetic and real data.

\subsection{Preliminaries}
\label{subsec:preliminaries}

\myparagraph{Latent diffusion model.}
We implement our method based on Stable Diffusion~\cite{stable_diffusion}, a probabilistic generative model that learns to generate realistic images using a textual prompt. 
Given data $(x,c) \in \gD$, where $x$ is an image and $c$ is a caption describing $x$, the model learns a conditional distribution $p(x|c)$ by gradually denoising the Gaussian noise in the latent space. Given a pretrained encoder $E$ that encodes an image $x$ to a latent $z$, i.e. $z=E(x)$, the objective function is defined as:
\begin{equation}
    \min_{\theta} \,\, \mathbb{E}_{(x,c) \sim \gD, \, \epsilon \sim \gN(0,1), \, t} \, \left[\, \left\| \, \epsilon - \epsilon_{\theta} (z_t, \tau(c), t) \, \right\|_2^2 \,\right] \, ,
\end{equation}
where $t$ is a timestep, $z_t$ is a latent noised $t$ steps from the latent $z$, $\tau$ is a text encoder, and $\epsilon_{\theta}$ is a latent diffusion model. Intuitively, the parameters $\theta$ are trained to denoise the latent $z_t$, given a text prompt $c$ as conditional information. In the inference phase, a random noise vector $z_T$ is passed through the latent diffusion model $T$ times, along with the caption $c$, to get a denoised latent $z_0$. $z_0$ is then fed into a pretrained decoder $D$ to get an image $x'=D(z_0)$ for the text-to-image generation. 
\\

\myparagraph{Low-rank adaptation.}
The Low-Rank Adaption method (LoRA)~\cite{lora}, is a fine-tuning method to adapt a large pre-trained model to downstream tasks in a parameter-efficient manner. Given pre-trained model weights $\theta \in \mathbb{R}^{d \times k}$, LoRA introduces a new parameter $\delta \in \mathbb{R}^{d \times k} $ that is decomposed into two matrices, $\delta=BA$, where $B \in \mathbb{R}^{d \times r}$ and $A \in \mathbb{R}^{r \times k}$ with small LoRA rank $r$, $r \ll \min (d, k)$. The LoRA weights are added to the model weights to obtain the fine-tuned weights, i.e. $\theta^{\text{(ft)}} = \theta \!+ \delta$, for adaptation to downstream tasks. During training, $\theta$ remains fixed while only $\delta$ is updated.

\begin{figure}[t]
    \centering
    \includegraphics[width=\linewidth]{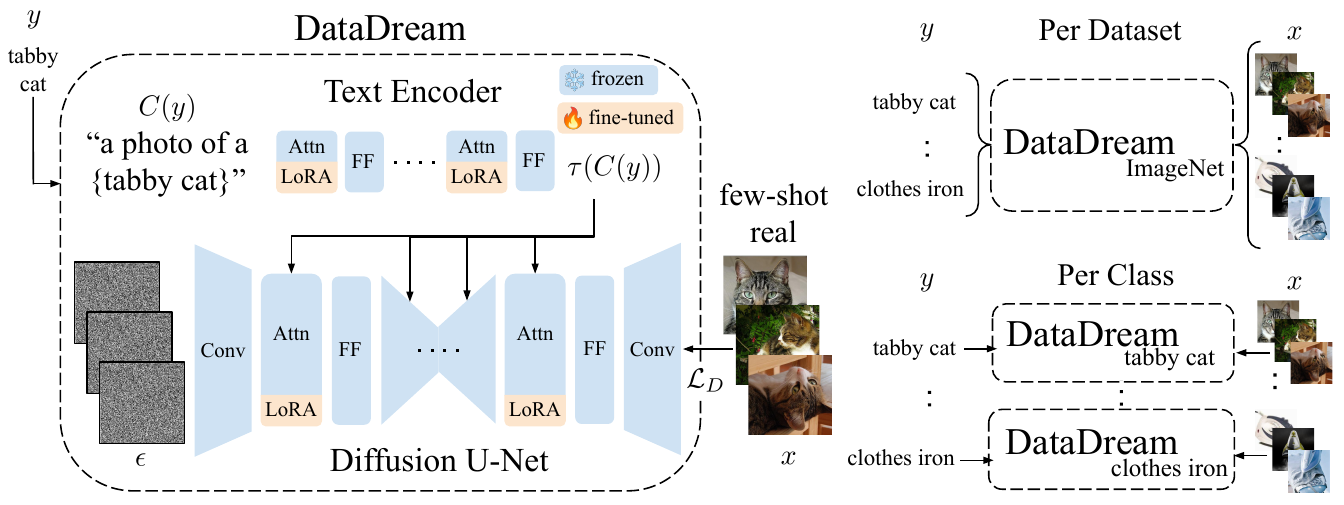} 
    \vspace{-10pt}
    \caption{Overview of DataDream. We fine-tune LoRA weights for the linear weights of the attention layers in both the text-encoder and the diffusion U-net to generate images closer to the few-shot images. We can train one set of DataDream weights for the whole dataset sharing common dataset-specific characteristics between classes, or a separate set of weights for each class to better learn fine-grained details of each classes. 
    }
    \label{fig:main}
\end{figure}

\subsection{\ours\ method}
\label{subsec:datadream}

Our goal is to improve classification performance by leveraging synthetic images generated by diffusion models. To this end, it is crucial to align the synthetic image distribution to that of the real images. We achieve alignment by adapting the diffusion model to a few-shot dataset of real images.

We assume access to a few-shot dataset $\gD^{\text{fs}}=\{(x_i, y_i)\}_{i=1}^{KN}$, where $x_i$ is an image, $y_i \in \{1,2,\cdots\!, N\}$ is its label, $K$ is the number of samples per class, and $N$ is the number of classes. To match the real data distribution, we fine-tune it with the few-shot dataset $\gD^{\text{fs}}$. 
Concretely, we introduce LoRA weights in both the text-encoder and the U-net of the diffusion model, where we make the parameter-efficient choice of adapting the attention layers.
For every attention layer, we consider the query, key, value, and output projection matrices $W_q, W_k, W_v, W_o$, where for each matrix, the linear projection is replaced by
\begin{equation}
    h_{l,\star} = W_{\star} h_{l-1} + B_{\star} A_{\star} h_{l-1}
\end{equation}
with $h$ representing the input/output activations of the projections, and resulting in the trainable LoRA weights $\delta^{(l)} = \{A_{\star}, B_{\star} | \forall \star \in \{q, k, v, o\}\}$ for every attention layer $l$. We omit bias weights for notational simplicity. All other model parameters are kept frozen (including $W_{\star}$) while $\delta$ weights are optimized with gradient descent. To start training from the pre-trained diffusion model checkpoint, weight matrices $B_{\star}$ are initialized with zeros while $A_{\star}$ is initialized randomly. As a result, the combined fine-tuning weights $B_{\star} A_{\star}$ are zero initially and incrementally learn modifications to the original pre-trained weights. At test time, LoRA weights can be integrated into the model by updating the weights with $W^{\text{(ft)}}_{\star} =W_{\star} + B_{\star} A_{\star}$, such that inference time is equivalent to the pre-trained model. In contrast to DreamBooth~\cite{dreambooth}, we do not fine-tune all network weights and do not add a preservation loss, as its regularization would prevent a strong alignment with the real images.

We further consider two settings: 1) \textbf{\oursdset}, where we train the LoRA weights of the diffusion model on the whole dataset $\gD^{\text{fs}}$, and 2) \linebreak \textbf{\ourscls}, where we initialize $N$ sets of LoRA weights $\{\delta_n|n=1,\cdots\!,N\}$, one for each of the dataset classes trained on the subset $\gD^{\text{fs}}_{n} = \{(x,y)| (x,y) \!\in \gD^{\text{fs}}, y\!=\!n\}$.

In the \oursdset\ setting, the original model parameters $\theta$ are kept frozen and only the LoRA weights are trained with the objective function
\begin{equation}
    \min_{\delta} \mathcal{L}_{\text{D}} = \min_{\delta} \,\, \mathbb{E}_{(x,y) \sim \gD^{\text{fs}}, \, \epsilon \sim \gN(0,1), \, t} \, \left[\, || \, \epsilon - \epsilon_{\theta\!, \delta} (z_t, \tau_{\delta}(C(y)), t) \, ||_2^2 \,\right] \, . 
    \label{eq:datadream_loss}
\end{equation}
In the \ourscls\ setting, $\gD^{\text{fs}}_{n}$ and $\delta_n$ would replace $\gD^{\text{fs}}$ and $\delta$, respectively. Since we use a text-to-image diffusion model, we define the text condition through the function $C$ which maps the label $y$, i.e. class name, to a prompt using the standard template, \texttt{"a photo of a [CLS]"}~\cite{clip,coop}. The prompt is passed through the text encoder and subsequently used during the decoding steps of the diffusion model. We illustrate both \ours\ fine-tuning and our two settings in Figure~\ref{fig:main}. 

Both settings have distinct advantages. In \oursdset, LoRA weight sharing between classes allows knowledge transfer about common characteristics within the whole dataset. 
This would be beneficial in a fine-grained dataset that shares the coarse-grained features across classes.
On the other hand, \ourscls\ allocates more weights to learn about 
details of each class, which allows the generation model to better align with the per-class data distribution. 

After adapting the diffusion model to the few-shot dataset, we generate 500 images per class with the adapted model conditioned on the same textual prompt used for \ours, forming a synthetic dataset $\gD^{synth}$. We train a classifier on either only synthetic images or the combination of synthetic and real few-shot images $\gD^{fs}$.

For classifier training, we adapt a CLIP model~\cite{clip}, similar to previous work in few-shot classification~\cite{disef}. We add LoRA adaptors~\cite{lora} to both image encoder and text encoder of CLIP ViT-B/16 model~\cite{clip}. When training with synthetic and real images jointly, we use a weighted average of the losses from real data and synthetic data,
\begin{equation}
    \mathcal{L}_{\text{C}} = \,\, 
    \lambda \, \mathbb{E}_{(x,y) \sim \gD^{\text{fs}}} \, \text{CE}(f(x),y) + 
    (1 \!-\! \lambda) \, \mathbb{E}_{(x,y) \sim \gD^{\text{synth}}} \, \text{CE}(f(x),y) \, ,
\end{equation}
where $\lambda$ is the weight assigned to the loss from real data and the function $\text{CE}$ is a cross-entropy loss.

\section{Experiments}
\label{sec:experiments}

In this section, we present our experimental results on \ours. We present details of the experimental setup in \S\ref{subsec:exp_setup}. In \S\ref{subsec:main_result}, we compare our methods to baselines both quantitatively and qualitatively. Furthermore, we analyze the synthetic data of \ours\ to understand why it outperforms baselines in \S\ref{subsec:analysis}, followed by ablation studies in \S\ref{subsec:ablation}.

\subsection{Experimental setup}
\label{subsec:exp_setup}

\myparagraph{Benchmarks.}
We evaluate our method on 10 datasets: ImageNet~\cite{imagenet}, Oxford Pets~\cite{pets}
containing fine-grained pet classes,
FGVC Aircraft~\cite{aircraft}
containing fine-grained aircraft classes,
Food101~\cite{food101}
containing common food classes,
Stanford Cars~\cite{cars}
containing fine-grained car classes,
DTD \cite{dtd}
with texture images,
EuroSAT~\cite{eurosat}
with satellite images,
Flowers 102~\cite{flowers102}
containing fine-grained flower classes,
SUN397~\cite{sun397}
with scene images,
and Caltech 101~\cite{caltech101} 
with pictures of common objects. \\

\myparagraph{Implementation details.} We implement \ours\ based on Stable Diffusion~\cite{stable_diffusion} version 2.1. For each seed, we randomly sample the few-shot images from the training samples of each dataset. Our method is trained for 200 epochs with a batch size of 8 for all datasets, with the exception of \oursdset\ on ImageNet, which is trained for 100 epochs. Hence, \oursdset\ and \ourscls\ share the same amount of training compute, i.e. each of the $N$ \ourscls\ adapter weights (one for each class) performs $S/N$ update steps where $S$ is the total number of steps of \oursdset\ for the whole dataset. We use AdamW~\cite{adamw} as an optimizer and learning rate $1e\!-\!4$, with a cosine annealing scheduler. We use LoRA rank $r\!=\!16$ for all adapted weights in \ours. For synthetic image generation with \ours, we use 50 steps and guidance scale 2.0. We generate 500 images per class if not mentioned otherwise. For the classifier, we use CLIP ViT-B/16~\cite{clip} as a base model, and fine-tune LoRA applied on both the image encoder and text encoder of CLIP with rank 16. We set the weight assigned to the real loss term to $\lambda\!=\!0.8$. \ours\ is computed on three random seeds. Additional implementation details can be found in Appendix~\ref{sec:app:details_dd}. \\

\myparagraph{Baseline methods.}
As all methods adapt CLIP ViT-B/16 as the classifier, we provide CLIP zero-shot performance as a baseline. In our first setting, we update the classifier using only synthetic data. For this, we compare against two alternative data-generation methods: IsSynth \cite{issynth} and DISEF \cite{disef}. In our second setting, classifier adaptation uses the synthetic data in addition to the few-shot real data. We refer to LoRA \cite{lora} training with only the real few-shot data as \oursbase, to signify that this is the baseline version of \ours, without the benefit of our synthetic data, as done in~\cite{disef}. \oursdset, \ourscls, and DISEF \cite{disef} all build upon this foundation. These experiments highlight the benefit of adding synthetic data to the real few-shot images. We also compare against several SOTA few-shot methods. In these, we include two Parameter Efficient Fine-Tuning (PEFT) techniques: VPT \cite{vpt} and CoOp \cite{coop}, which only use real few-shot data. We additionally compare to two SOTA image generation techniques, IsSynth \cite{issynth} and DISEF \cite{disef}. For fair comparisons, we use Stable Diffusion v2.1 to generate images for all baselines instead of the originally used GLIDE~\cite{nichol2021glide} or Stable Diffusion v1.5. More details are described in Appendix~\ref{sec:app:details_baseline}.

\subsection{Classification performance with \ours}
\label{subsec:main_result}

\begin{table*}[t]
\centering
\setlength{\tabcolsep}{2pt}
\resizebox{\textwidth}{!}{
\renewcommand{\arraystretch}{1.4}
\begin{tabular}{lcc|cccccccccc|c}
\toprule
\bf Method & R & S & IN & CAL & DTD & EuSAT & AirC & Pets & Cars & SUN & Food & FLO & Avg \\ \hline \hline
CLIP (zero-shot)~\cite{clip} & & &  
70.2 & 96.1 & 46.1 & 38.1 & 23.8 & 91.0 & 63.1 & 63.8 & 85.1 & 71.8 & 64.1 \\ \hline \hline
IsSynth~\cite{issynth} &  & \vmark & 
70.0$_{\pm0.6}$ & 95.7$_{\pm0.7}$ & 67.6$_{\pm0.5}$ & 71.3$_{\pm2.9}$ & 34.5$_{\pm3.4}$ & 92.1$_{\pm0.6}$ & 65.9$_{\pm0.0}$ & 72.2$_{\pm0.0}$ & 85.4$_{\pm0.1}$ & 90.0$_{\pm0.2}$ & 74.5$_{\pm0.9}$ \\
 DISEF~\cite{disef} &  & \vmark & 
67.1$_{\pm0.2}$ & 93.4$_{\pm1.0}$ & 66.1$_{\pm0.6}$ & 69.2$_{\pm2.7}$ & 26.8$_{\pm2.2}$ & 91.0$_{\pm0.0}$ & 63.2$_{\pm0.0}$ & 73.5$_{\pm0.1}$ & 85.1$_{\pm0.0}$ & 85.4$_{\pm0.7}$ & 72.1$_{\pm0.8}$ \\
\ourscls\ (ours) &  & \vmark & 
\bf 71.6$_{\pm0.2}$ & \bf 96.4$_{\pm0.0}$ & 68.6$_{\pm2.0}$ & \bf 85.4$_{\pm2.9}$ & 60.3$_{\pm0.9}$ & \bf 94.2$_{\pm0.4}$ & 90.5$_{\pm0.3}$ & 74.5$_{\pm0.0}$ & \bf 86.9$_{\pm0.1}$ & 97.2$_{\pm0.2}$ & 82.6$_{\pm0.7}$ \\ 
\oursdset\ (ours) & & \vmark & 
71.5$_{\pm0.0}$ & 96.2$_{\pm0.1}$ & \bf 69.5$_{\pm1.2}$ & 80.3$_{\pm4.1}$ & \bf 71.2$_{\pm0.1}$ & 94.0$_{\pm0.1}$ & \bf 92.2$_{\pm0.1}$ & \bf 74.5$_{\pm0.1}$ & 86.7$_{\pm0.1}$ & \bf 98.0$_{\pm0.4}$ & \bf 83.4$_{\pm0.7}$ \\ \hline \hline
\oursbase & \vmark & & 
73.4$_{\pm0.2}$ & 96.8$_{\pm0.1}$ & 78.3$_{\pm2.8}$ & 93.5$_{\pm0.7}$ & 59.3$_{\pm2.8}$ & 94.0$_{\pm0.1}$ & 87.5$_{\pm0.6}$ & 77.1$_{\pm0.1}$ & \bf 87.6$_{\pm0.0}$ & 98.7$_{\pm0.1}$ & 84.6$_{\pm0.8}$ \\
IsSynth~\cite{issynth}  & \vmark & \vmark & 
73.9$_{\pm0.1}$ & 97.4$_{\pm0.2}$ & \bf 81.6$_{\pm0.4}$ & 93.9$_{\pm0.1}$ & 64.8$_{\pm0.8}$ & 92.1$_{\pm0.1}$ & 88.5$_{\pm0.3}$ & \bf 77.7$_{\pm0.0}$ & 86.0$_{\pm0.0}$ & 99.0$_{\pm0.0}$ & 85.5$_{\pm0.2}$ \\
DISEF~\cite{disef}  & \vmark & \vmark & 
73.8$_{\pm0.2}$ & 97.0$_{\pm0.1}$ & 81.5$_{\pm0.6}$ & \bf 94.0$_{\pm0.5}$ & 64.3$_{\pm0.4}$ & 92.6$_{\pm1.2}$ & 87.9$_{\pm0.5}$ & 77.6$_{\pm0.1}$ & 86.2$_{\pm0.6}$ & 99.0$_{\pm0.2}$ & 85.4$_{\pm0.4}$ \\
\ourscls\ (ours) & \vmark & \vmark & 
73.8$_{\pm0.1}$  & \bf 97.6$_{\pm0.2}$ & \bf 81.6$_{\pm0.4}$ & 93.8$_{\pm0.3}$ & 68.3$_{\pm0.4}$ & 94.5$_{\pm0.3}$ & 91.2$_{\pm0.2}$ & 77.5$_{\pm0.1}$ & 87.5$_{\pm0.1}$ & \bf 99.4$_{\pm0.2}$ & 86.5$_{\pm0.4}$\\
\oursdset\ (ours) & \vmark & \vmark & 
\bf 74.1$_{\pm0.3}$ & 96.9$_{\pm0.7}$ & \bf 81.6$_{\pm0.6}$ & 93.4$_{\pm0.0}$ & \bf 72.3$_{\pm0.2}$ & \bf 94.8$_{\pm0.3}$ & \bf 92.4$_{\pm0.1}$ &  77.5$_{\pm0.1}$ & \bf 87.6$_{\pm0.1}$ & \bf 99.4$_{\pm0.1}$ & \bf 87.0$_{\pm0.4}$\\
\bottomrule
\end{tabular}
}
\vspace{5pt}
\caption{
    \textbf{Few-shot classification performance with \ours\ using real 16-shot and synthetic images} where the training dataset includes synthetic data only (top), or synthetic data + 16 real shots (bottom). All results use CLIP ViT-B/16 as the base classification model, and 500 synthetic images generated by 16 real shots. Datasets are IN: ImageNet, CAL: Caltech 101, EuSAT: EuroSAT, AirC: FGVC Aircraft, FLO: Flowers 102. R/S means using real/synthetic images for fine-tuning. \ours\ and baseline methods are computed on three random seeds.}
\label{tbl:quant_pooled}
\vspace{-10pt}
\label{table:main}
\end{table*}

\myparagraph{Quantitative results on solely synthetic data.}
We refer to the upper portion of Table \ref{tbl:quant_pooled} for the synthetic-only setting, where we show that \ours-generated data achieves state-of-the-art results on all 10 datasets. For example on FGVC Aircraft \cite{aircraft}, \oursdset\ achieves an impressive 47.4\% point increase over the CLIP zero-shot model. In addition, on Stanford Cars \cite{cars} \oursdset\ achieves 92.2\%, while IsSynth \cite{issynth} is at 65.9\% and DISEF \cite{disef} at 63.2\%. On Flowers102 \cite{flowers102}, \oursdset\ obtains 98.0\% while IsSynth and DISEF reach only 90.0\% and 85.4\%, respectively. These boosts signify that \ours\ is able to closely follow the real few-shot data distribution in its generated images.

We believe that this evaluation benchmark allows the best assessment of the quality of the synthetic image generations for training image classifiers. While adding real data to the synthetic images at training time generally provides a performance boost, it also makes it harder to quantify the quality of the synthetic images for the task, because most of the improvement still stems from the real images. Hence, issues with synthetic data generation, such as redundancy or class misrepresentation, will be more visible in synthetic-only benchmarks.

Comparing \ourscls\ and \oursdset, the results are split over method superiority. We hypothesize that this difference comes from inherent dataset properties. For datasets where all classes share high visual similarity, sharing weights becomes beneficial as distribution characteristics generalize across classes. For example, we find that FGVC Aircraft \cite{aircraft} and Stanford Cars \cite{cars} show a significant advantage of \oursdset\ over \ourscls. On the other hand, datasets where classes span a wide range benefit from fully specializing to the unique classes, as seen in the results for Caltech101 \cite{caltech101} and Food101 \cite{food101}. \\

\myparagraph{Quantitative results on real $+$ synthetic data.}
In Table \ref{tbl:quant_pooled} (bottom), we present the results for the synthetic + real setting. \oursbase\ provides the foundation for this section, consisting of LoRA applied to CLIP with the real few-shot data. DISEF, \ourscls, and \oursdset\ build upon this by adding their respective synthetic images to the training data, which is generated from the same few-shot data. Comparing \oursdset\ and \ourscls\ to \oursbase, we observe that our synthetic data improves performance on 9 out of 10 datasets over naive use of real few-shot data. For example, on FGVC Aircraft \cite{aircraft}, our synthetic data facilitates an improvement of 13.0\% over using the real few-shot data naively. In comparison, DISEF only achieves a 5.0\% increase. Furthermore, we improved upon Stanford Cars \cite{cars} by 4.9\%, where DISEF saw only a 0.4\% increase. This shows that generating images with \ours\ consistently provides value not only over naive use of the few-shot examples, but also over other data generation techniques. In fact, especially in case of the Stanford Cars dataset, the real images do not provide more information than the synthetic images generated by our model (92.2\% on synthetic only vs 92.4\% real + synthetic settings), which is an exciting observation. \\

\begin{table*}[t]
\centering
\setlength{\tabcolsep}{2pt}
\resizebox{\textwidth}{!}{
\renewcommand{\arraystretch}{1.4}
\begin{tabular}{lc|cccccccccc|c}
\toprule
\bf Method & S & IN & CAL & DTD & EuSAT & AirC & Pets & Cars & SUN & Food & FLO & Avg \\ \hline \hline
VPT~\cite{vpt} & & 
69.6 & 95.4 & 66.1 & 92.3 & 36.2 & 91.8 & 69.0 & 70.5 & 87.0 & 91.0 & 76.9\\
CoOp~\cite{coop} & & 
68.0 & 95.2 & 70.7 & 87.1 & 45.5 & 89.9 & 81.4 & 73.0 & 83.7 & 97.6 & 79.2\\
IsSynth~\cite{issynth} & \vmark & 
73.9 & \bf 97.4 & \bf 81.6 & 93.9 & 64.8 & 92.1 & 88.5 & \bf 77.7 & 86.0 & 99.0 & 85.5 \\ 
DISEF~\cite{disef}  & \vmark & 
73.8 & 97.0 & 81.5 & \bf 94.0 & 64.3 & 92.6 & 87.9 & 77.6 & 86.2 & 99.0 & 85.4 \\ 
\oursdset\ (ours) & \vmark & 
\bf 74.1 & 96.9 & \bf 81.6 & 93.4 & \bf 72.3 & \bf 94.8 & \bf 92.4 & 77.5 & \bf 87.6 & \bf 99.4 & \bf 87.0\\
\bottomrule
\end{tabular}
}
\vspace{5pt}
\caption{
    \textbf{Comparing \ours\ with few-shot SOTA.} We compare \ours\ with SOTA few-shot methods. The base setting, dataset abbreviations, and setting notations match those in Table \ref{tbl:quant_pooled}. S indicates methods using synthetic data generation.
}
\label{tbl:quant_sota}
\vspace{-10pt}
\end{table*}

\myparagraph{Quantitative results comparing with SOTA.} We compare \ours\ with SOTA few-shot methods in Table \ref{tbl:quant_sota}. Ours, i.e. \oursdset, improves over the previous SOTA on 7 out of 10 datasets, while being competitive on the other 3. On the FGVC Aircraft \cite{aircraft} dataset, we improve SOTA by 7.5\%, on Pets \cite{pets} by 2.2\%, and on Stanford Cars \cite{cars} by 3.9\%. This highlights that synthetic images generated by \ours\ provide more training value than the previous SOTA generation method. On average, we improve over the next best synthetic augmentation method by 1.5\% and over the best method without data generation, CoOp~\cite{coop}, by 7.8\%.

\subsection{Analysis of \ours}
\label{subsec:analysis}

\begin{figure}[t]
    \centering
    \includegraphics[width=\textwidth]{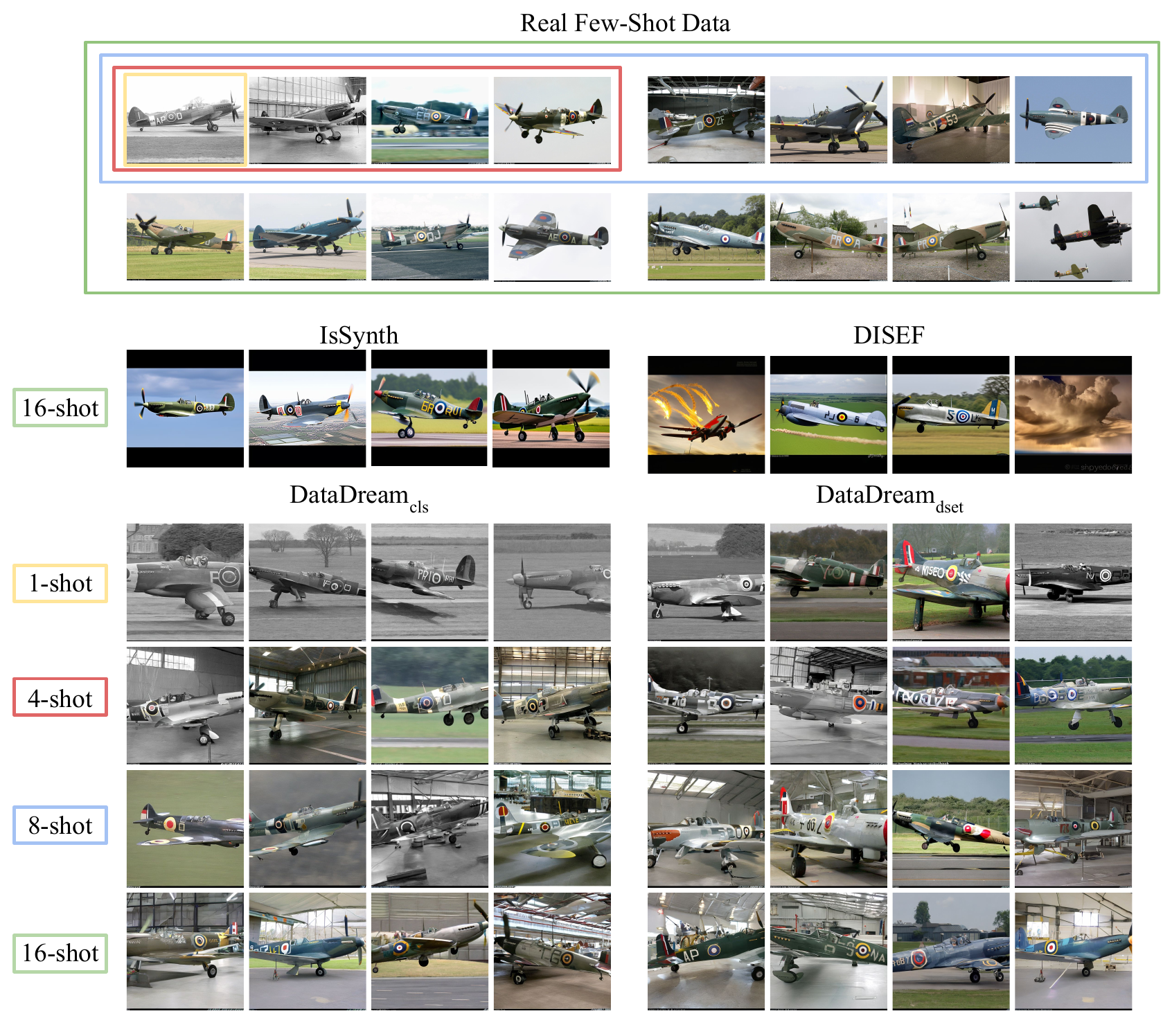} 
    \caption{\textbf{Qualitative results with increasing number of shots vs 16-shot images generated with SOTA} of the class Spitfire from the FGVC Aircraft \cite{aircraft} dataset. The real few-shot images at the top are used to generate the presented synthetic images at the bottom. We always use a fixed set of 16 samples, i.e. 1-shot image is a subset of 16-shots, to insure fairness in comparing results with the increasing number of shots.}
    \vspace{-10pt}
    \label{fig:qualitative}
\end{figure}

\myparagraph{Qualitative results.} 
We provide a qualitative analysis of \ours\ in Figure~\ref{fig:qualitative} of the Spitfire class in FGVC Aircraft. To support our 1-, 4-, 8-, and 16-shot generated images, we include the real few-shot examples used to generate them. We also show previous SOTA images for comparison, from two other image-generation methods: DISEF \cite{disef} and IsSynth \cite{issynth}.

When comparing to the previous SOTA, we notice that \ours\ is better able to generate images that match the target domain. For example, they imitate that in the context of the dataset, planes are more likely to be photographed on the ground, in a hangar, or taking off than in the air, unlike both previous methods that generate images that are unlikely to be found in the dataset. We also notice that our models are better able to match the color palette of the real data, as opposed to DISEF, where we find the colors to be too bright compared to the real data. Furthermore, our model is the only one of all three that replicates the black border at the bottom of all images, even after only a single shot.

Furthermore, we notice that DISEF has a higher tendency to generate out-of-distribution data, sometimes omitting the target class entirely and therefore creating a need for CLIP filtering. We hypothesize this might be due to their use of diverse captions, which may sometimes guide the image generation too far from the core distribution. By focusing on fidelity to the class distribution and keeping our prompts simple, we generate fewer out-of-distribution samples. This allows us to use all samples generated, which is a better use of resources.

We also notice some interesting differences between \oursdset\ and \newline \ourscls. On the one hand, we find that \ourscls\ is better able to accurately represent the Spitfire class, especially at a low number of shots. On the other hand, the additional data in \oursdset\ allows it to avoid certain overfitting mistakes, such as creating only black and white images after the first shot, which happens to be a monochrome image.

Comparing \ours\ models trained on different numbers of few-shot examples, we notice an increase in quality with number of images, showing qualitatively the benefit of adding even a few more images. Already at four shots, we obtain images that are not only better quality, but closer to the real data domain. We also note that the lower the number of shots, the more the model benefits from careful selection of a diverse and representative group, so that the model does not pick up on patterns that are not representative of the full data distribution. At only four shots, we notice that the real images contain a specific color palette that is not necessarily representative of the full dataset, as evidenced by the next four images; this led to the 4-shot results lacking diversity of color and brightness. This goes to show that wherever possible, careful selection of representative data is highly beneficial. It also highlights the ability of our method to find and replicate patterns in the few-shot distribution. \\

\begin{figure}[t]
    \centering
    \includegraphics[width=0.6\linewidth]{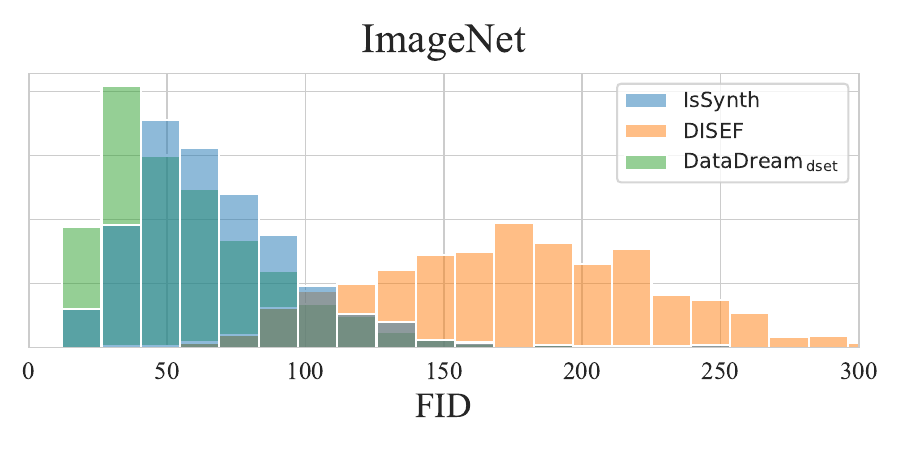} 
    \vspace{-10pt}
    \caption{
        \textbf{Distribution of FID scores per-class.}        
        The FID score is calculated per-class to measure how close the synthetic data distribution is to the real data distribution. 
    }
    \label{fig:fid_per_class}
\end{figure}

\myparagraph{Distribution alignment.}
The qualitative analysis shows \ours\ being able to capture both the presence of objects of interest and the fine-grained features essential for class discrimination. 
To gain more insights, we examine the alignment between synthetic and real datasets. To quantitatively assess the alignment, we use the Frechet Inception Distance (FID)~\cite{fid} score, a metric that quantifies the quality of generated images. Concretely, we compute the set of FID scores for each method by evaluating the distance between the distribution of synthetic images and that of real images on a per-class basis. Lower FID indicates synthetic images are closer to the real data distribution. We visualize the FID scores using histograms, as shown in Figure~\ref{fig:fid_per_class}.

In the experiment on ImageNet, we observe that the histogram for our method skews left, indicating lower FID scores. Meanwhile, the histogram of DISEF tends to lean towards the right. This is attributed to how DISEF uses LLM-generated prompts for image generation. While it gives the generated images diversity, it also introduces out-of-distribution artifacts, as observed in the qualitative analysis. Compared to DISEF, IsSynth aligns more with the real data, but may have less diversity due to its usage of a standard prompt, generating similar images from the conditioned real image. In contrast, our \oursdset\ balances fidelity, due to the adaptation of the generative model to match few real shots, and diversity, due to the initial randomness in the generation pipeline. This results in the synthetic images of \oursdset\ closely matching the real data distribution. We posit that the better alignment contributes to classification performance, as demonstrated in \S\ref{subsec:main_result}. 

\subsection{Ablation study}
\label{subsec:ablation}

\begin{figure}[t]
    \centering
    \begin{minipage}{0.49\textwidth}
        \centering
        \includegraphics[width=\linewidth]{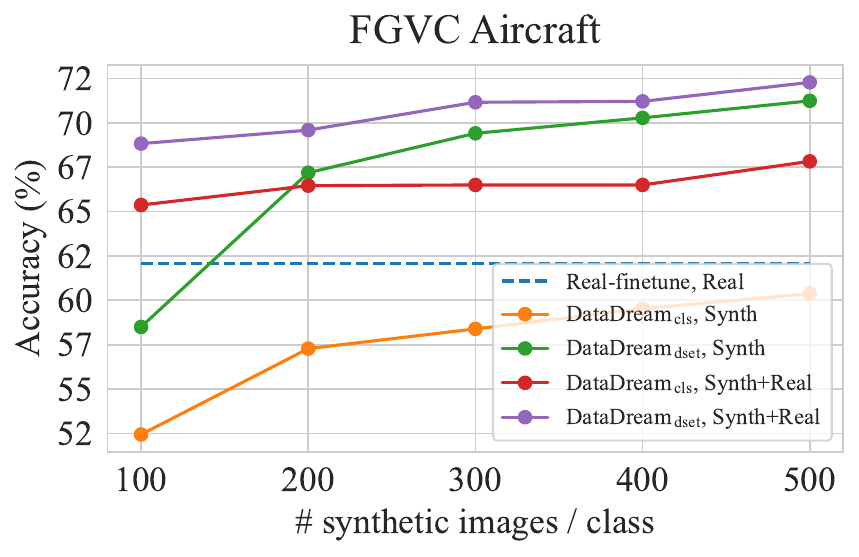}
    \end{minipage}\hfill
    \begin{minipage}{0.49\textwidth}
        \centering
        \includegraphics[width=\linewidth]{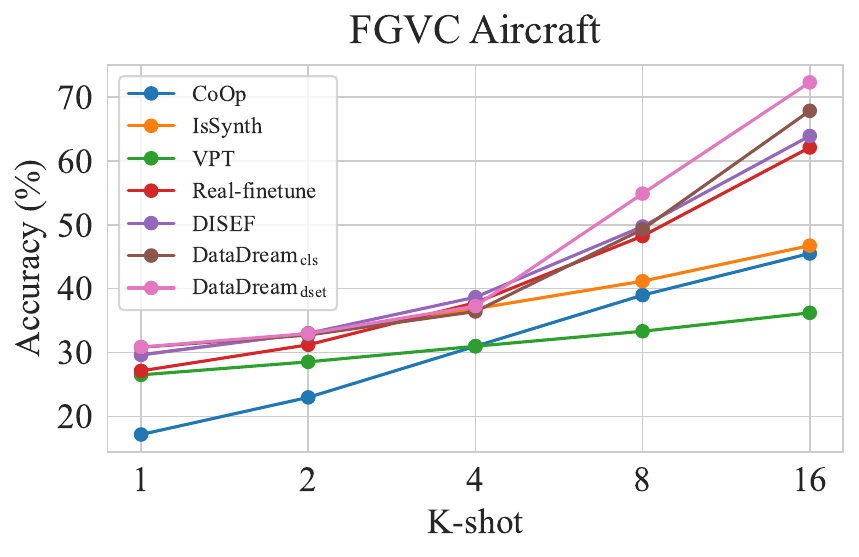}
    \end{minipage}
    \caption{\textbf{Ablation study of \ours.} Left: We vary the number of synthetic images per class to understand the scaling effect. Right: We vary the number of real examples used for training \ours. 
    }    
    \label{fig:ablation}
\end{figure}

\myparagraph{Accuracy scaling by number of synthetic images.} 
Previous literature has shown that as the number of synthetic images increases, model accuracy may also increase \cite{fakeit}. Therefore, we provide Figure~\ref{fig:ablation}, where the left part shows the effect on \ours\ of increasing the number of images for FGVC Aircraft \cite{aircraft}. We find that as more images are generated, model accuracy increases in all settings: \oursdset\ and \ourscls\, and Synth and Synth $+$ Real. Even at 500 images, we observe that performance is not yet saturated. Compared to \oursbase, we observe that in the Synth $+$ Real setting, \ours\ performs better already starting at 100 images per class.

In the synthetic-only setting, \ourscls\ out-performs real images entirely after generating only 200 images. Another interesting result is the gap between \ourscls\ and \oursdset: this difference holds between any number of images, showing that \oursdset\ is a better fit for this dataset than \ourscls, regardless of the number of synthetic images. However, remembering from Table \ref{tbl:quant_pooled} that \oursdset\ performed better than \ourscls\ on 5 out of 10 datasets, we note that this advantage is dataset-dependent rather than a general trend. \\

\myparagraph{Varying the number of few-shot images $K$.}
Furthermore, we operate in a $K$-shot regime; therefore, we expect that as $K$ increases, the model accuracy should increase as well. As done in \cite{disef}, we show the effect of 1-, 2-, 4-, 8-, and 16-shots on FGVC Aircraft \cite{aircraft} dataset, for 500 images in the real + synth setting, compared to previous literature. We observe that as the number of few-shot images increases, \ours\ consistently shows higher accuracy than previous SOTA. We believe that this behavior is expected; as with any training or fine-tuning regime, at least a small training dataset base is necessary. Too few samples could be prone to overfitting, thus reducing variety and failing to include enough information to successfully understand the overall class distribution. At the same, since we use LoRA on a subset of all model parameters, we limit the amount of overfitting from our fine-tuning as compared to full model fine-tuning. This allows \ours\ to outperform \oursbase\ even when we use only a single shot. As more data becomes available, however, \ours\ is able to successfully leverage even as few as four or eight shots to noticeably adapt to the data distribution, as was shown in Section \ref{subsec:analysis}. Hence, we obtain a relative performance boost when compared to other methods.

\section{Conclusion}
\label{sec:conclusion}

In this paper, we studied the efficacy of leveraging the generative models for improving the image classification performance in few-shot scenarios. We proposed \ours, a method to generate synthetic data with the guidance of few-shot samples, which are then used for training the image classifier. We introduced LoRA adaptors on both the text encoder and the diffusion U-Net to efficiently fine-tune the generative model. We proposed two variants: \oursdset, which trains LoRA on the whole targeted dataset, and \ourscls, which adapts LoRA per-class. Our experiments demonstrate that our method consistently improves classification performance across benchmarks, both in the synthetic-only and synthetic+real settings. Through qualitative analysis, we observed that images generated by our method more precisely generate objects of interest as well as fine-grained details, contributing to their alignment with real data distributions, as quantitatively examined by FID scores. Furthermore, we investigated the scalability of our method by increasing the number of synthetic data samples and the number of real samples.

\newpage


\section*{Acknowledgements}

Jae Myung Kim thanks the International Max Planck Research School for Intelligent Systems (IMPRS-IS)
and the European Laboratory for Learning and Intelligent Systems (ELLIS) PhD programs for support. This work was supported by the ERC (853489 - DEXIM) and the BMBF (Tübingen AI Center, FKZ: 01IS18039A). 
Cordelia Schmid would like to acknowledge the support by the K\"orber European Science Prize. Also, the authors gratefully acknowledge the Gauss Centre for Supercomputing e.V. (www.gauss-centre.eu) for funding this project by providing computing time on the GCS Supercomputer JUWELS at Jülich Supercomputing Centre (JSC). 

%
%
\bibliographystyle{splncs04}
\bibliography{main}

\newpage
\appendix
\section*{\centering Supplementary Material for \\DataDream: Few-shot Guided Dataset Generation}
\renewcommand\thesection{\Alph{section}}

\section{Broader Impacts and Limitations}

While DataDream primarily focuses on generating synthetic image datasets for image classification, our approach can be extended to other domains and tasks. For example, sentiment classification can benefit from synthetic datasets generated by large language models fine-tuned on a few challenging samples. Additionally, DataDream can be applied to other image modalities. For instance, the scarcity of medical data often impedes the performance of medical AI. Our method could serve as an augmentation tool to enhance performance in such scenarios.

However, it's important to recognize the limitations of our approach. We observe, in EuroSAT and DTD, a performance gap when comparing models trained solely on DataDream synthetic datasets to those trained solely on real few-shot data, with the latter performing better. We speculate that this disparity comes from the challenges generative models face when learning from out-of-distribution data. It becomes difficult to fine-tune generative models with few-shot samples that are different from the data the models were originally trained on. For instance, satellite land images, those in the EuroSAT dataset, may not be well-represented in the LAION dataset used for training Stable Diffusion models. Consequently, the diffusion model struggles to generalize and accurately interpret satellite images, even after fine-tuning with real few-shot satellite images.

Moreover, by training on synthetic training, our method inherits the limitations of the underlying generative models. For instance, generative models will propagate biases that appear in the training data, such as social or gender biases. As a result, classifiers trained on the synthetic data of a generative models are also prone to carrying forward these biases. Especially when employing proprietary generative models, it is often unknown what data they were trained on.

DataDream fine-tunes the generative model using few-shot samples. Because the dataset is much smaller, it is easier to control for biases that could potentially be introduced in this stage, but this would require manual intervention when curating the few-shot dataset.
At the same time, malicious actors could purposefully introduce biases through the DataDream fine-tuning, for example to construct a classifier that discriminates against minorities. Detecting biases in the resulting classifier might then become more difficult than observing them from generative images.
We believe that investigating and mitigating bias in synthetic data generation would be an important area for future research.

\section{Implementation details}
\label{sec:app:details_dd}

Following the methodology of DISEF~\cite{disef} for the classifier training, we consider different learning rates, weight decay, and whether to use Mixup~\cite{mixup} and Cutmix~\cite{cutmix} methods for data augmentation as a hyper-parameter. 
Concretely, we use batch size 64, and AdamW~\cite{adamw} as an optimizer with a cosine annealing scheduler. The learning rate is searched in $\{1e\!-\!4, 1e\!-\!5, 1e\!-\!6, 1e\!-\!7\}$ and the weight decay in $\{5e\!-\!4, 1e\!-\!4\}$. We use standard augmentation methods as default, i.e. random resized crop, random horizontal flip, random color jitter, and random gray scale, while searching on whether to additionally use Mixup and Cutmix. We set the weight assigned to the real loss term as $\lambda\!=\!0.8$.

\section{Baseline methods}
\label{sec:app:details_baseline}

We compare our \ours\ method to two state-of-the-art image generation methods in the few-shot setting, IsSynth~\cite{issynth} and DISEF~\cite{disef}. While these methods originally use GLIDE~\cite{nichol2021glide} and Stable Diffusion v1.5~\cite{stable_diffusion}, respectively, for the generative model, we use the Stable Diffusion v2.1 and follow the other pipelines suggested in each paper. We implement both methods based on DISEF official code\footnote{https://github.com/vturrisi/disef}. When running the IsSynth method, we replace the textual prompt conditioned on the diffusion model from the caption generated by LLaVA~\cite{llava} to a standard prompt, e.g. \texttt{"a photo of a [CLS]"}, as suggested in IsSynth. When we compare \ours\ to IsSynth and DISEF in Table~\ref{tbl:quant_pooled}, we use the same generation procedure of generating 500 images without filtering out them but using them all for training the classifier.

\begin{table}[t]
\renewcommand\thetable{B}
\centering
\setlength{\tabcolsep}{2pt}
\resizebox{0.8\textwidth}{!}{
\renewcommand{\arraystretch}{1.035}
\begin{tabular}{lcc|cccc|cccc|cccc}
\toprule
\multicolumn{3}{c|}{} & \multicolumn{4}{|c|}{[A]} & \multicolumn{4}{|c|}{\small CLIP ViT-L/14} & \multicolumn{4}{|c}{CLIP RN50} \\
\bf  & R & S & AirC & Cars & Food & CAL & AirC & Cars & Food & CAL  & AirC & Cars & Food & CAL\\ \hline \hline
%
IsSynth [13] & & \vmark & 
24.09 & 69.59 & 84.79 & 94.58 & 
30.52 & 81.64 & 90.18 & 97.01 &  
27.22 & 74.42 & 48.78 & 87.33\\ 
DISEF [9] & & \vmark & 
26.03 & 63.82 & 84.86 & 93.90 & 
31.59 & 71.55 & 90.20 & 96.40 &  
22.25 & 33.36 & 46.03 & 83.70 \\ 
\ourscls & & \vmark & 
39.97 & 79.90 & 85.44 & \bf 94.63 & 
69.21 & 94.21 & \bf 91.72 & 97.78 &   
67.79 & 92.31 & 61.03 & 90.05 \\ 
\oursdset & & \vmark & 
\bf 43.75 & \bf 81.37 & \bf 85.20 & 94.62 & 
\bf 76.33 & \bf 94.53 & 91.51 & \bf 97.94 &  
\bf 75.31 & \bf 93.34 & \bf 64.16 & \bf 91.22 \\ \hline 
base (fewshot) & \vmark & & 
40.61 & 75.12 & 79.05 & 92.55 & 
72.01 & 91.18 & 91.92 & \bf 98.46 & 
61.57 & 78.86 & 63.52 & 93.29 \\ 
IsSynth [13] & \vmark & \vmark & 
43.10 & 80.50 & 86.30 & 95.48 & 
73.33 & 93.68 & 91.96 & 98.1 & 
70.94 & 90.82 & 68.77 & 94.54 \\ 
DISEF [9] & \vmark & \vmark & 
42.62 & 79.41 & 85.97 & 95.40 & 
73.83 & 92.67 & 91.94 & 98.30 & 
65.99 & 79.18 & \bf 70.10 & 94.34 \\ 
\ourscls & \vmark & \vmark & 
48.41 & 83.78 & 86.67 & 95.47 & 
77.69 & 94.71 & \bf 92.16 & 98.22 &  
79.21 & 92.99 & 66.70 & 94.37 \\ 
\oursdset & \vmark & \vmark & 
\bf 49.31 & \bf 83.87 & \bf 86.71 & \bf 95.62 & 
\bf 80.98 & \bf 95.18 & 92.14 & 98.30 & 
\bf 81.46 & \bf 93.30 & 66.63 & \bf 94.62\\ 
\bottomrule
\end{tabular}
}
\vspace{5pt}
\caption{Compatibility with other CLIP fine-tuning and classifiers.}
\label{tbl:app:other_models}
\end{table}

\section{Compatibility with other CLIP fine-tuning and classifiers}
DataDream generates data for downstream tasks, so it is compatible with other CLIP fine-tuning methods. We evaluate the synthetic dataset on the most recent suggested method for training the CLIP model \cite{wang2024a}. Moreover, we evaluate our and other datasets with two additional classifiers: CLIP ViT-L/14 and CLIP RN50. For both, we fully fine-tune the visual encoder and LoRA fine-tune the text encoder.  
As shown in the table~\ref{tbl:app:other_models}, DataDream works better than the baselines in both CLIP fine-tuning method~\cite{wang2024a} and other classifiers for both sythetic only and real + synthetic settings.


\section{Ablation study}

\begin{wraptable}{r}{0.45\textwidth}
\vspace{-23pt}
\centering
\setlength{\tabcolsep}{2pt}
\resizebox{0.45\textwidth}{!}{
\renewcommand{\arraystretch}{1.4}
\begin{tabular}{lcc|cc}
\toprule
\bf Method & txt enc. & w/o pre. loss & Synth & Real+Synth \\ \hline \hline
\oursdset\ & & & 
19.96 & 62.49 \\
\oursdset\ & \vmark & & 
18.58 & 62.61 \\
\oursdset\ & & \vmark & 
59.80 & 66.78 \\ \hline
\oursdset\ (ours) & \vmark & \vmark & 
\bf 71.07 & \bf 71.98 \\
\bottomrule
\end{tabular}
}
\vspace{-5pt}
\caption{\footnotesize{
    \textbf{Ablation study of using different pipeline for \ours\ training.}
    Mark on ``txt enc.'' indicates training LoRA on the text encoder in addition to the UNet (no mark = UNet only). Mark on ``w/o pre. loss'' indicates using preservation loss~\cite{dreambooth} in addition to the reconstruction loss (Equation~\ref{eq:datadream_loss}). 
}}
\label{tbl:sup:ablate_pipeline}
\vspace{-15pt}
\end{wraptable}
%
For our \ours\ method, we take inspiration from DreamBooth~\cite{dreambooth} which proposes to finetune a diffusion model to generate personalized images. To enable the generation of personalized images across diverse prompts beyond the one it was trained on, e.g. \texttt{"a photo of a [CLS]"}, DreamBooth introduces a preservation loss. This loss acts as a regularizer such that the fine-tuned model maintains its original capabilities when generating images in the absence of personalized tokens in the textual prompt. Since we are not interested in employing our fine-tuned model for general-purpose image generation, we put more focus on faithful replication of the few-shot data distribution than preserving the generation quality of irrelevant generations.

We conduct an ablation study to investigate the impact of preservation loss on data generation within \ours. Additionally, We explore the effect of applying LoRA solely to the UNet, or to both the UNet and text encoder of the diffusion model. We train each combination setting for 200 epochs on the FGVC Aircraft~\cite{aircraft} dataset, and the results are shown in Table~\ref{tbl:sup:ablate_pipeline}.

First, we observe that applying LoRA to both the UNet and text encoder (without preservation loss) improves the accuracy from 59.80\% to 71.07\% on the classifier trained solely with synthetic data, and 66.78\% to 71.98\% when incorporating real 16-shot data with synthetic data. This improvement indicates that additional fine-tuning of the text encoder enhances the model's ability to capture the features present in few-shot data. Second, we observe that using the preservation loss decreases the accuracy from 71.07\% to 18.58\% in the synthetic setting and 71.98\% to 62.61\% in the real+synth setting. This decline suggests that, the inclusion of preservation loss hinders the model's adaptation to the target few-shot data, limiting its generation performance. Furthermore, given that we employ the same standard prompt for both training \ours\ and generating images with the \ours-fintuned model, the necessity for the preservation loss for \ours\ training is diminished. Overall, our findings suggest that using LoRA adaptor in the text encoder and excluding the preservation loss achieve the best performance.

\section{$K$-shot varying $K$ on additional datasets}

\begin{figure}[]
    \centering
    \begin{subfigure}{0.45\textwidth}
        \includegraphics[width=\textwidth]{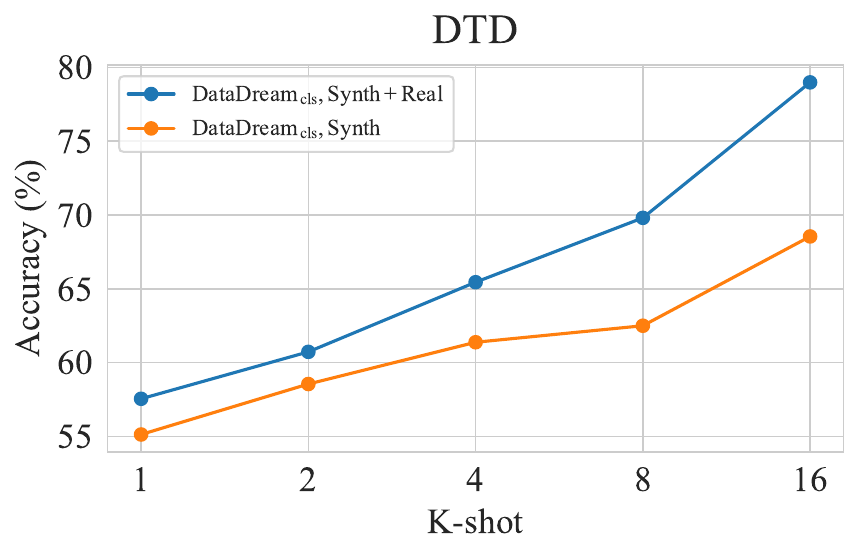} 
        \caption{\textbf{\ourscls\ accuracy scaling by number of shots for DTD} \cite{dtd}.}
        \label{fig:shot_vary_app_dtd}
    \end{subfigure}
    \begin{subfigure}{0.45\textwidth}
        \includegraphics[width=\textwidth]{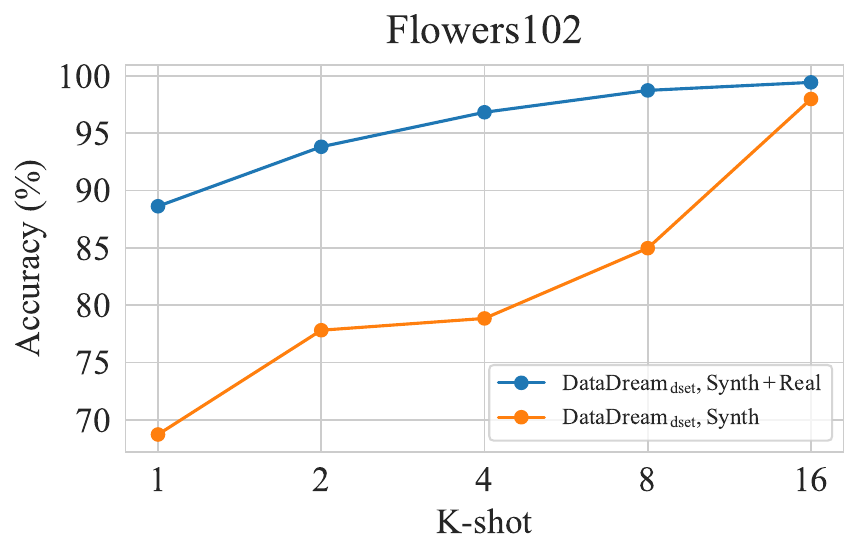} 
        \caption{\textbf{\oursdset\ accuracy scaling by number of shots for Flowers102} \cite{flowers102}.}
        \label{fig:shot_vary_app_flowers}
    \end{subfigure}
    \begin{subfigure}{0.45\textwidth}
        \includegraphics[width=\textwidth]{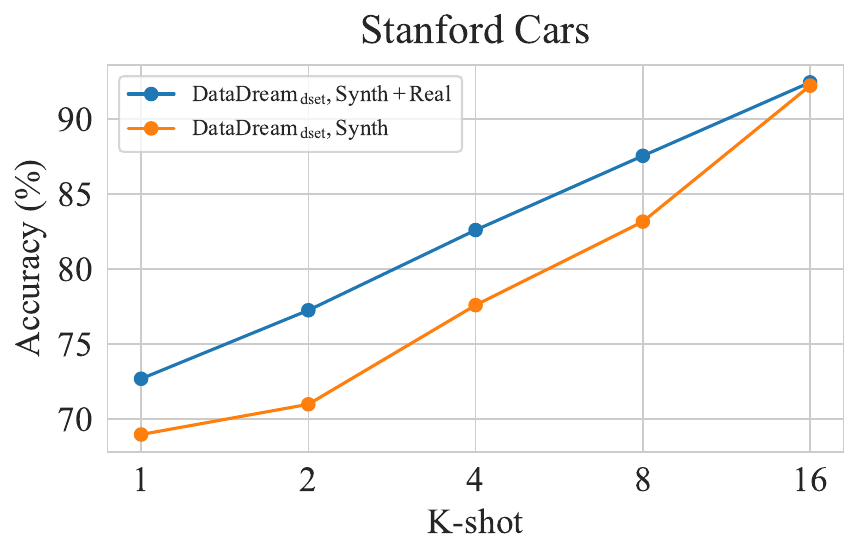} 
        \caption{\textbf{\oursdset\ accuracy scaling by number of shots for Stanford Cars} \cite{cars}.}
        \label{fig:shot_vary_app_cars}
    \end{subfigure}
    \begin{subfigure}{0.45\textwidth}
        \includegraphics[width=\textwidth]{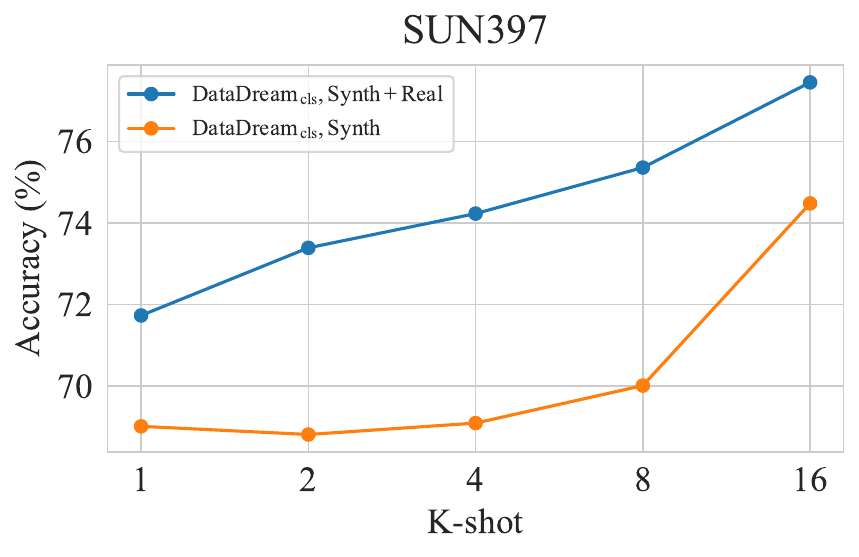} 
        \caption{\textbf{\ourscls\ accuracy scaling by number of shots for Sun397} \cite{sun397}.}
        \label{fig:shot_vary_app_sun}
    \end{subfigure}
    \vspace{-10pt}
    \label{fig:shot_vary_app}
\end{figure}
In this section, we include additional scaling graphs on four additional datasets, which show \ours\ performance improves as we increase $K$ in the $K$-shot setting. The included datasets are DTD \cite{dtd} (Figure \ref{fig:shot_vary_app_dtd}), Flowers102 \cite{flowers102} (Figure \ref{fig:shot_vary_app_flowers}), Stanford Cars \cite{cars} (Figure \ref{fig:shot_vary_app_cars}), and SUN397 \cite{sun397} (Figure \ref{fig:shot_vary_app_sun}). For each dataset, we show results with the better of our two models, as reported in Table \ref{tbl:quant_pooled}: \oursdset\ for Flowers102 and Stanford Cars, and \ourscls\ for DTD \cite{dtd} and SUN397 \cite{sun397}. We include results for both using only synthetic (Synth) as well as the combination of synthetic and real data (Synth+Real).

One aspect we would like to highlight is the performance similarity on Stanford Cars \cite{cars} between the Synth and Real+Synth settings on 16 shots. This can already be seen in Table \ref{tbl:quant_pooled}, but especially stands out in Figure \ref{fig:shot_vary_app_cars}. This means that by sixteen shots, \oursdset\ can faithfully represent the information from the real data, to the point where there is no performance gained from additionally training on the real samples.

\section{Qualitative examples}
We provide additional qualitative results for one further FGVC Aircraft class and two extra datasets. As done in Figure \ref{fig:qualitative}, we designate the real few-shot images provided for 1-, 4-, 8-, and 16-shots. We then show the 16-shot images generated by two competing methods, IsSynth \cite{issynth} and DISEF \cite{disef}. Finally, we have images generated by both \oursdset\ and \ourscls, for varying number of shots.

\begin{figure}[t]
    \centering
        \includegraphics[width=\textwidth]{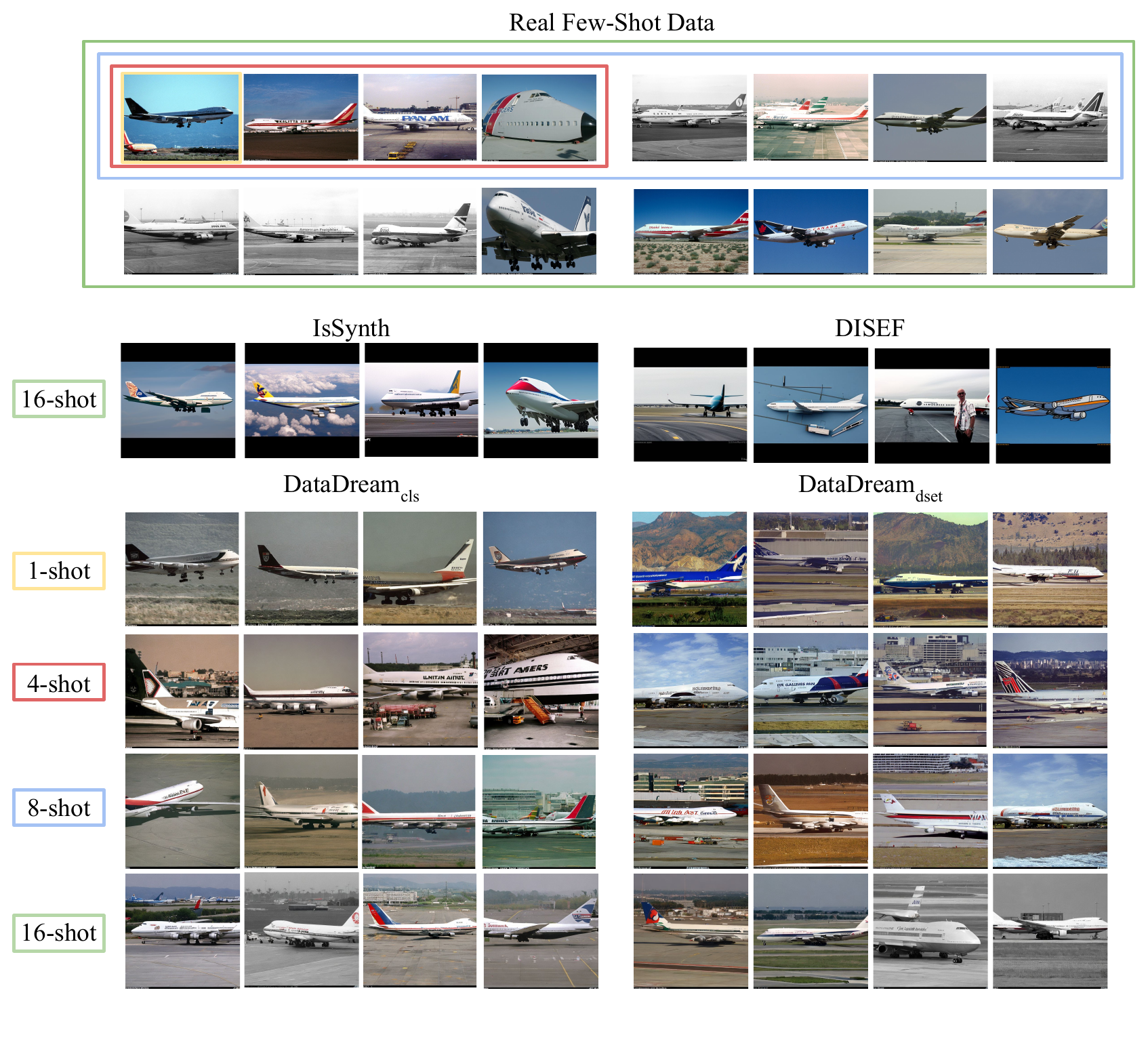} 
    \caption{\textbf{Qualitative results of the class 747-100 from the FGVC Aircraft} \cite{aircraft} dataset, created the same as Figure \ref{fig:qualitative}.}
    \vspace{-10pt}
    \label{fig:qualitative-747}
\end{figure}

In Figure \ref{fig:qualitative-747}, we see images from the class 747-100 in FGVC Aircraft \cite{aircraft}, which is a commercial aircraft variety. When comparing with previous SOTA, we once again see that \ours\ generally provides better in-distribution data. 
DISEF in particular generates many incorrect modalities (e.g. cartoon or toy) or images with irrelevant primary subjects. In comparison, \ours\ consistently generates images with the correct shape, as seen from angles commonly found in the dataset.

\begin{figure}[t]
    \centering
        \includegraphics[width=\textwidth]{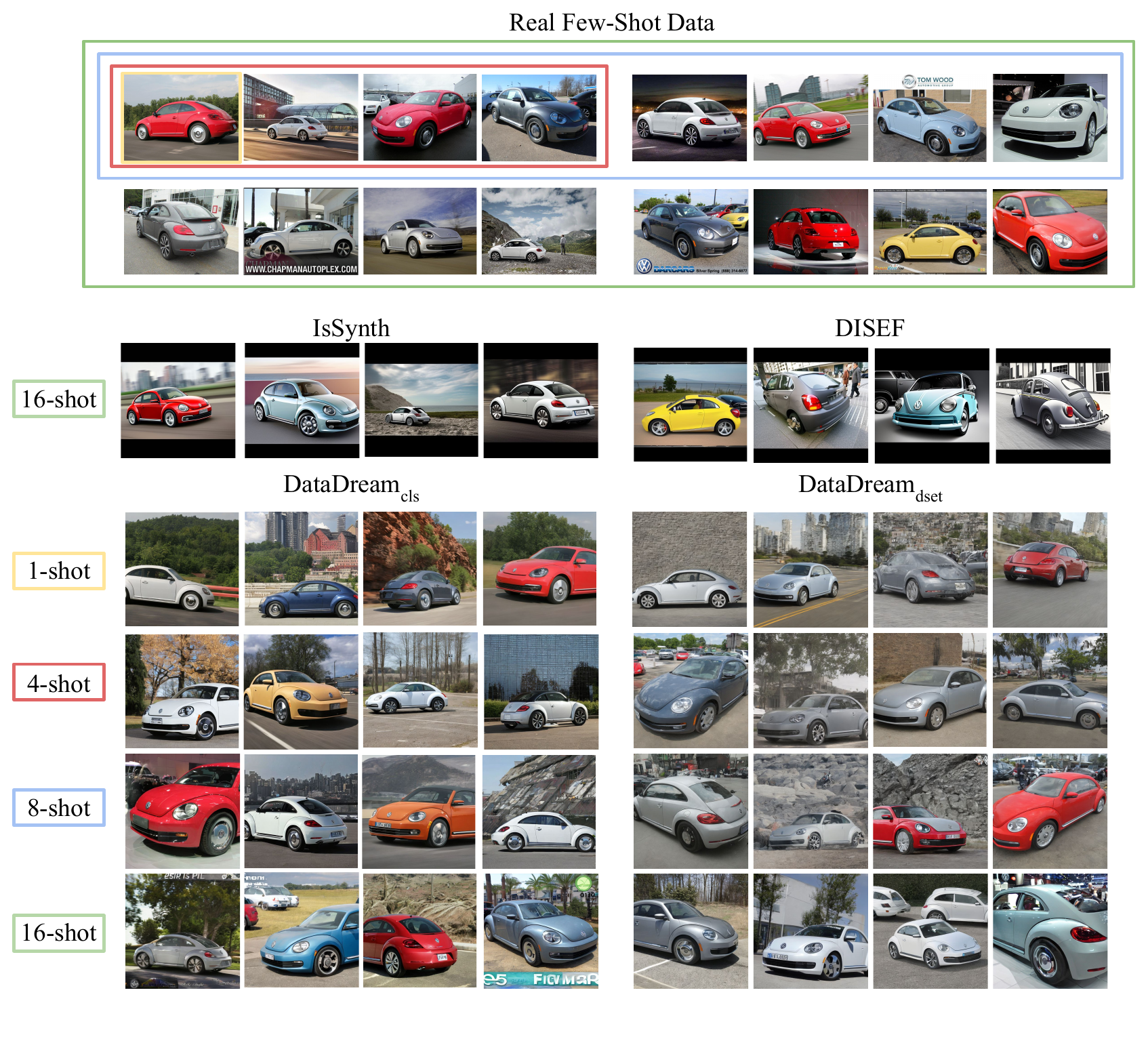} 
    \caption{\textbf{Qualitative results of the class Volkswagen Beetle HatchBack 2012 from the Stanford Cars} \cite{cars} dataset, created the same as Figure \ref{fig:qualitative}.}
    \vspace{-10pt}
    \label{fig:qualitative3-bug}
\end{figure}

In Figure \ref{fig:qualitative3-bug}, we show images of Stanford Cars' \cite{cars} Volkswagen Beetle Hatchback 2012 class. Once again, we see that \ours\ is better at consistently generating the correct car than DISEF, and generates backgrounds closer to what is found in the real dataset than IsSynth. Furthermore, we would like to point out a coloring difference between \oursdset\ and \ourscls. We know that Volkswagen Beetles are available in a wide variety of colors; \ourscls\ demonstrates this by generating cars in yellow, orange, multiple shades of bright blue, etc. On the other hand, most car varieties are available in a smaller color pool, many of which are muted. We see that \oursdset-generated cars are more likely to be white, gray, or red, all of which are colors commonly found in other cars. There are still several shades of blue generated, but they are more muted than those generated by \ourscls. Hence, we see an example of how \oursdset\ can learn patterns from the wider dataset and apply them to individual classes where it may not be optimal, while \ourscls\ can differentiate better. We remember from Table \ref{tbl:quant_pooled} that \oursdset\ performed better than \ourscls\ on the full dataset, so overall, the value of information shared between classes was greater than what was lost. 

Finally, we provide examples of generated images from the Sword Lily class in the Flowers102 dataset \cite{flowers102}. First, we notice that while both previous methods struggle to generate faithful representations of the class, \ourscls\ generates accurate images from a single shot, and \oursdset\ from four shots. One interesting aspect is the number of flowers per image. At a single and four shots, \ourscls\ generates a few images of single or double flowers. However, by four-shots for \ourscls\ and from the first shot for \oursdset, flowers are almost always generated in bunches, outside. While this is representative of the majority of images, there are also examples in the real dataset with a low number of flowers. Hence, there may still be more to gain in terms of ensuring that the entire distribution is represented in the synthetic dataset. We leave this for future work to explore.

\begin{figure}[t]
    \centering
        \includegraphics[width=\textwidth]{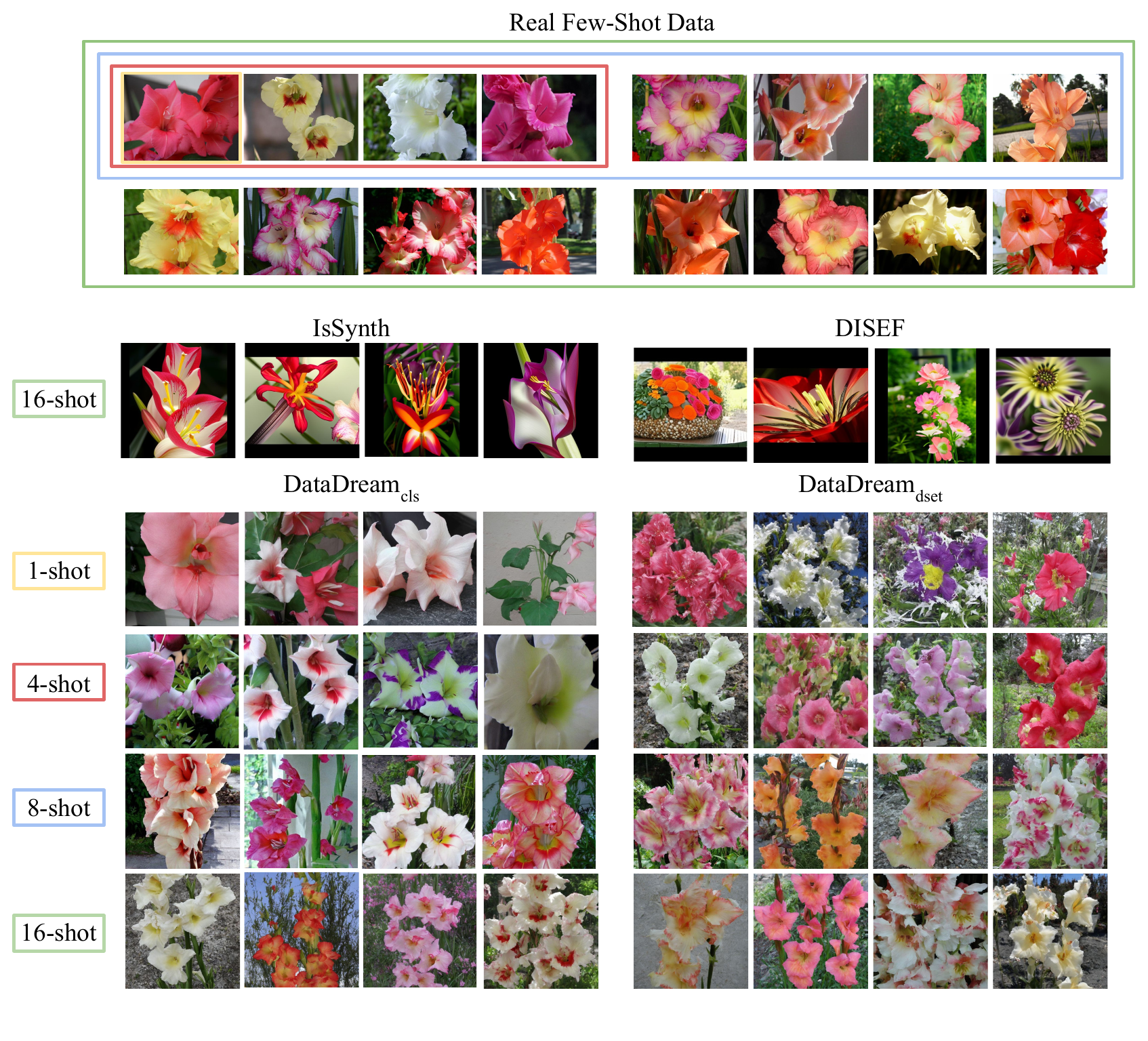} 
    \caption{\textbf{Qualitative results of the class Sword Lily from the Flowers102} \cite{flowers102} dataset, created the same as Figure \ref{fig:qualitative}.}
    \vspace{-10pt}
    \label{fig:qualitative-swordlily}
\end{figure}

\end{document}